\documentclass{article}

\usepackage{microtype}
\usepackage{graphicx}
\usepackage{subfigure}
\usepackage{booktabs} 

\usepackage{hyperref}

\usepackage[accepted]{icml2024}

\usepackage{interval}
\usepackage{amsmath}
\usepackage{amssymb}
\usepackage{mathtools}
\usepackage{amsthm}

\usepackage[capitalize,noabbrev]{cleveref}

\usepackage{float}
\usepackage{caption} 

\usepackage{multirow}
\usepackage{multicol}
\usepackage{xcolor}
\usepackage{colortbl}
\usepackage{placeins}
\newcommand{\highlightrow}{\rowcolor{teal!15}}

\theoremstyle{plain}

\theoremstyle{definition}

\theoremstyle{remark}

\usepackage[textsize=tiny]{todonotes}

\icmltitlerunning{Characterizing Prompt Compression Methods for Long Context Inference}

\begin{document}

\twocolumn[
\icmltitle{Characterizing Prompt Compression Methods for Long Context Inference}




\begin{icmlauthorlist}
\icmlauthor{Siddharth Jha}{berkeley}
\icmlauthor{Lutfi Eren Erdogan}{berkeley}
\icmlauthor{Sehoon Kim}{berkeley}
\icmlauthor{Kurt Keutzer}{berkeley}
\icmlauthor{Amir Gholami}{berkeley}
\end{icmlauthorlist}

\icmlaffiliation{berkeley}{UC Berkeley}

\icmlcorrespondingauthor{Siddharth Jha}{sidjha@berkeley.edu}

\icmlkeywords{Machine Learning, ICML}

\vskip 0.3in
]



\printAffiliationsAndNotice{} 
\begin{abstract}

Long context inference presents challenges at the system level with increased compute and memory requirements, as well as from an accuracy perspective in being able to reason over long contexts.
Recently, several methods have been proposed to compress the prompt to reduce the context length.
However, there has been little work on comparing the different proposed methods across different tasks through a standardized analysis. This has led to conflicting results.
To address this, here we perform a comprehensive characterization and evaluation of different prompt compression methods.
In particular, we analyze extractive compression, summarization-based abstractive compression, and token pruning methods.
Surprisingly, we find that extractive compression often outperforms all the other approaches, and enables up to $10\times$ compression with minimal accuracy degradation.
Interestingly, we also find that despite several recent claims, token pruning methods often lag behind extractive
compression. We only found marginal improvements on summarization tasks. 
\end{abstract}
\section{Introduction}
\begin{figure}[!t]
    \centering
    \includegraphics[width=0.9\linewidth]{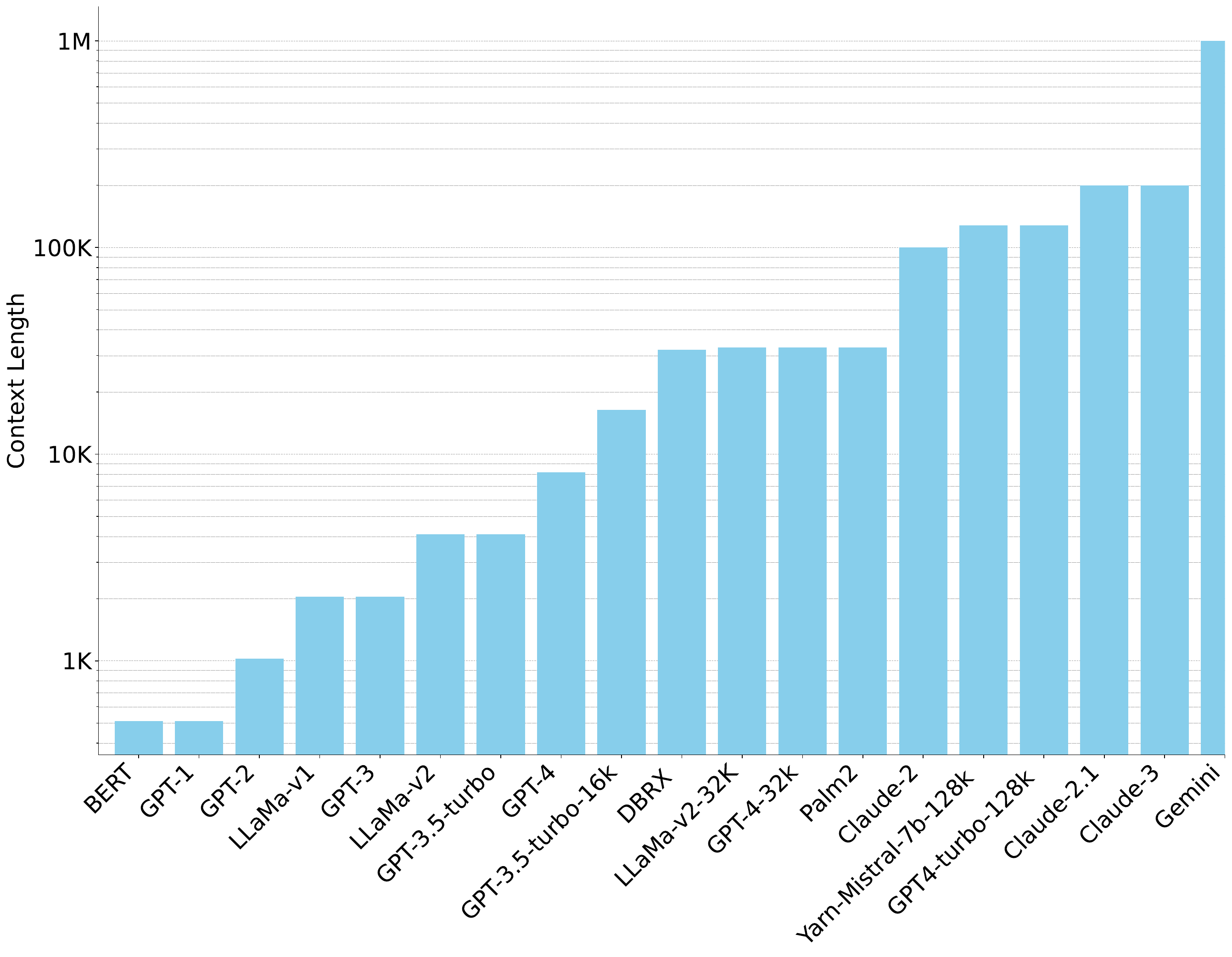}
    \caption{LLM context length has been rapidly increasing as many applications
    can benefit from longer context lengths.
    However, this often comes with accuracy challenges as LLMs seem to struggle
    with reasoning over long context lengths, along with higher cost and time to first token.
    }
    \label{fig:context_length_scaling}
\end{figure}

In recent years, the use of LLMs has experienced exponential growth, leading to a surge in applications that manage extensive textual contexts. 
For example, OpenAI's flagship GPT-3/3.5-Turbo/4-Turbo models have been exponentially increasing in context window size from a few thousand tokens to 128K tokens and Google Gemini model has publicly available models that support up to 1M context length (see~\autoref{fig:context_length_scaling}). 
The ability to perform long context inference is crucial in fields like legal and financial document analysis, copilots for large code bases~\cite{wu2023bloomberggpt,yang2023fingpt}, summarization~\cite{xiao2019extractive}, and interactive systems maintaining ongoing dialogues~\cite{packer2023memgpt}.
However, building applications that support long prompts presents significant system-level challenges, including increased computational demands, memory requirements, and costs~\cite{hooper2024kvquant, kim2023stack}. 
There is also the potential for a decline in the model's reasoning capabilities over extended sequences~\cite{liu2024lost}. 
Consequently, numerous prompt compression methods have been proposed, which aim to condense prompt lengths while preserving essential information.
Despite growing interest in prompt compression techniques, little is known about the behavior of such techniques due to a lack of standardized analysis, making it challenging for practitioners to choose the appropriate method for different applications.
For example, certain methods evaluate on context sizes of tens of thousands of tokens, while others on only a few hundred. 
Apart from initial context length, the evaluated compression rates and tasks also greatly vary.

To address these challenges, we perform a comprehensive characterization and evaluation of different prompt compression methods. In particular:

\begin{itemize}
    \item 
    We characterize methods into extractive compression, abstractive compression, or token pruning. We further distinguish methods as being query-agnostic or query-aware.
    Then we perform a comprehensive survey on each of these classes (see~\autoref{sec:related-work-prompt-compression}).
    \vspace{-2mm}
    \item We evaluate each paradigm on three single-document QA, multi-document QA, and summarization datasets. Furthermore, we study the impact of chunk size, query-aware abstractive summarization, and other important choices when building prompt compression systems (see~\autoref{sec:additional-analysis}).
    \vspace{-2mm}
    \item 
    Surprisingly, we find that extractive compression often outperforms all the other approaches, and enables up to $10\times$ compression with minimal accuracy degradation. Interestingly, we also find that despite several recent claims, token pruning methods often lag behind extractive compression. We only found marginal improvements on summarization tasks (see~\autoref{sec:experiments} and~\autoref{fig:gpt-3.5-pareto}).
\end{itemize}

\section{Related Work}
\subsection{Long Context LLMs}
\label{sec:long-conntext-llms}
There has been significant growth in context windows of LLMs in recent years. 
For example, Google's Gemini~\cite{reid2024gemini} supports context windows of up to 1M tokens in its publicly available API. 
Anthrtopic's Claude 3 models support context windows of 200k tokens~\cite{anthropic}, and OpenAI's GPT-4-Turbo model supports 128k tokens~\cite{openai}. 
Long prompts are naturally occurring in a variety of applications, such as those performing summarization, processing legal and financial documents~\cite{wu2023bloomberggpt, yang2023fingpt}, and chat agents which store prior conversation histories~\cite{packer2023memgpt}. 
However, there are a variety of challenges when using long context models. 
From the systems perspective, compute and memory requirements of the attention operator scale quadratically with sequence length. 
This has motivated researchers to explore a variety of techniques such as sparsity~\cite{zhang2024h2o,ge2023model,li2024snapkv} and quantization~\cite{hooper2024kvquant,liu2024kivi} to increase long context system efficiency. For those relying on LLM API providers, long prompts may lead to prohibitively expensive expenditure. Furthermore, the reasoning ability of language models has been shown to decrease at large prompt lengths~\cite{liu2024lost}. 
This is due to a lost in the middle effect where relevant context is not properly utilized when in the middle of a large context window.

\subsection{Retrieval-Augmented Generation}
Retrieval-augmented generation (RAG) is increasingly utilized in knowledge-intensive LLM applications to enhance performance by incorporating relevant external information into the model's decision-making process~\cite{lewis2020retrieval}. 
Typically, this is done by first breaking a large text corpus of relevant information into smaller chunks, with each chunk then embedded by an embedding model~\cite{gautier2022unsupervised,wang2022text}. 
To find relevant context for a user question, the question is also embedded and then similarity search is performed on the chunk embeddings to retrieve the most similar chunks. 
An important decision to make is determining how many chunks to retrieve. 
Retrieving too few chunks risks missing key information and retrieving too many chunks leads to long prompt sizes, which comes with the challenges mentioned in~\autoref{sec:long-conntext-llms}. 
Furthermore, certain applications may be using models without long context windows, in which case prompting the model with many chunks becomes impossible. 
From a cost, latency, and accuracy perspective, it is optimal to provide the minimum amount of information required to answer the question. This has motivated a series of prompt compression methodologies~\cite{ali2024prompt, jiang2023llmlingua, jiang2023longllmlingua, xu2023recomp}.

\subsection{Prompt Compression}
\label{sec:related-work-prompt-compression}
\begin{figure*}[!ht]
    \centering
    \includegraphics[width=\linewidth]{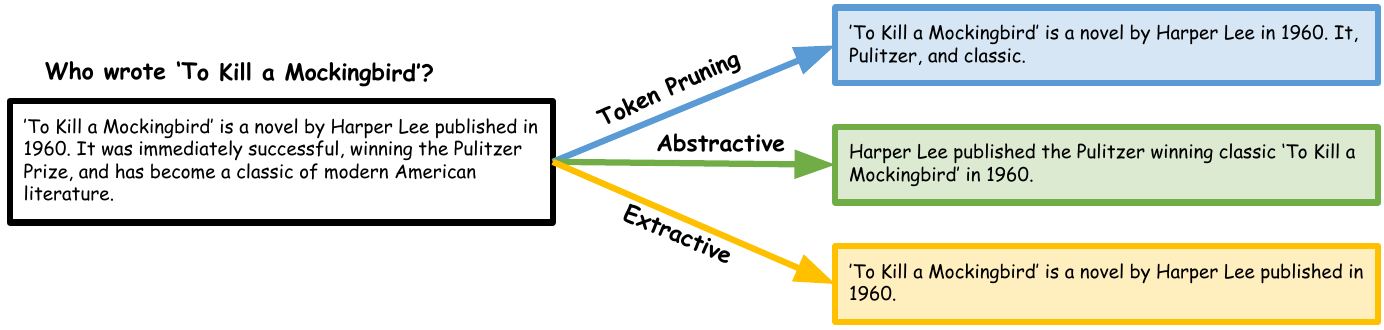}
    \caption{An illustration of different prompt compression methods. \textbf{Token pruning} methods like LongLLMLingua~\cite{pan2024llmlingua2}, Selective-Context~\cite{li2023compressing}, and PCRL~\cite{jung2023discrete} perform compression by discarding irrelevant tokens. \textbf{Abstractive compression} methods like Prompt-SAW~\cite{ali2024prompt}, RECOMP, and PRCA~\cite{yang2023prca} generate summaries by synthesizing information. \textbf{Extractive compression} methods like RECOMP~\cite{xu2023recomp} and reranker-based compression select documents, sentences, or phrases from the original context without altering them. In this example, each of the methods compresses the original context while keeping the necessary information to determine the book's author.}
    \label{fig:methods}
\end{figure*}

Prompt compression is the process of taking a long prompt and distilling only the most critical information in order to minimize length while still retaining necessary information. This can be done by either directly manipulating the text or by manipulating text embeddings. 
As an example of the latter, LLoCO~\cite{tan2024lloco} uses an encoder model to produce token embeddings from the original context. 
These token embeddings are then fed as the compressed context to a fine-tuned decoder model. Similar approaches are used in~\cite{chevalier2023adapting, ge2023context}.
While embedding-based compression methods show strong compression performance, such methods require extensive fine-tuning and significant changes to the inference pipeline, thereby restricting their application on API-based LLM services (e.g. OpenAI models).
Therefore, our main focus in this paper is on direct text manipulation as it requires minimal changes to the inference pipeline and can be used with LLM API providers.
Overall, existing text-based prompt compression methods can largely be categorized into three buckets: token pruning, abstractive compression, and extractive compression. We show an illustration of each paradigm in~\autoref{fig:methods}.

\subsubsection{Token Pruning Based Compression}
Token pruning methods perform compression by discarding irrelevant tokens. 
Selective-Context~\cite{li2023compressing} uses a small language model to judge self-information of tokens.
Then, tokens with low self-information are pruned from the original prompt.
LLMLingua~\cite{jiang2023llmlingua} is a similar method to Selective-Context but uses perplexity to determine the importance of tokens. 
LLMLingua first performs coarse-grained pruning by removing entire in-context examples and then performs fine-grained token pruning on the prompt.
LongLLMLingua~\cite{jiang2023longllmlingua} is a modification of LLMLingua designed for long context prompt compression.
Unlike LLMLingua, LongLLMLingua considers the perplexity of the question when conditioned on supporting documents to determine which documents are most relevant. 
After performing coarse-grained compression by removing irrelevant documents, fine-grained token pruning is performed by considering the perplexity of tokens before and after being conditioned on the question. The drop in perplexity after conditioning on the question is used to judge the relevance of a token. Tokens with low relevance are pruned.
PCRL~\cite{jung2023discrete} uses reinforcement learning to train a policy network to remove tokens in the original context. 
Specifically, the state seen by the policy is the original context and the action taken by the policy is a binary string denoting whether each token in the original context is kept or removed. 
The ROUGE~\cite{lin2004rouge} between the output from the original context and the output from the compressed context is considered as the reward to maximize.
The policy network is a frozen pre-trained small language model (e.g. GPT-2) a MLP head for binary classification. 
There has also been extensive research on token pruning methods for white-box Transformer models~\cite{goyal2020power,kim2020length,kim2022learned, wang2021spatten}. 
Such methods utilize the Transformer model's attention map at each layer in order to determine which tokens are least attended to by other tokens. 
These tokens are pruned before the sequence proceeds to the next layer in the Transformer. 
For the purposes of black-box prompt compression, a smaller white-box model may be used for token pruning, with the unpruned tokens from the white-box model being sent to the black-box LLM.

\begin{figure*}[!t]
    \centering
    \includegraphics[width=\linewidth]{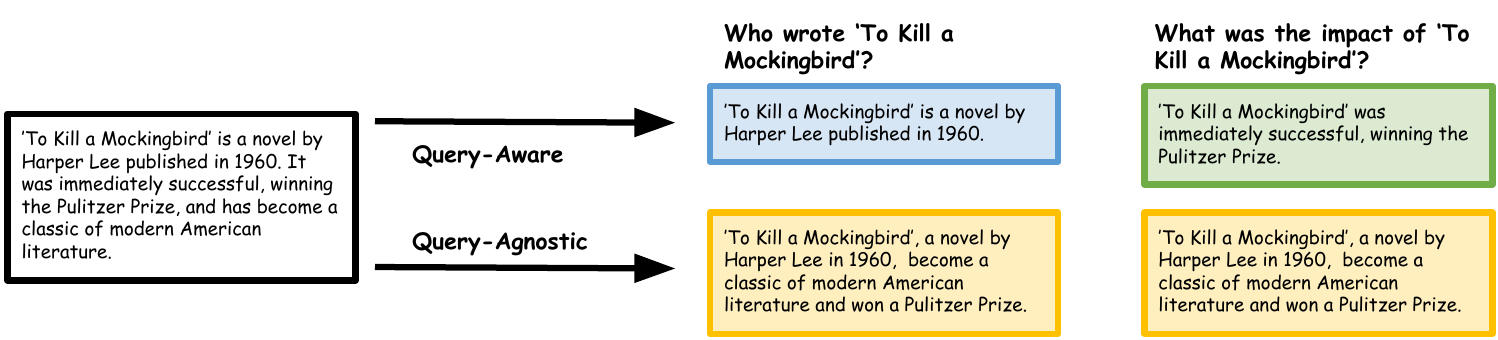}
    \caption{An illustration of \textbf{query-aware} and \textbf{query-agnostic} compression applied to a document in the prompt context.  With query-aware compression, the compressed context of the document changes based on the user's specific query, presenting a tailored version each time. Conversely, query-agnostic compression maintains a consistent compressed context of the document, irrespective of the query presented.}
    \label{fig:query-aware-agnostic}
\end{figure*}

\subsubsection{Abstractive Compression}
Abstractive compression techniques rely on summarization techniques to reduce the length of the original context.
RECOMP's~\cite{xu2023recomp} abstractive compressor is a fine-tuned T5-Large (775M) model~\cite{raffel2020exploring} that summarizes the initial context into a more compact form. 
By prompting the summarizer with the question at inference time, they generate query-aware summaries.
In the fine-tuning training data, they drive the summarization model to produce an empty string if a summarized context leads to performance degradation on the downstream task. 
To omit the fine-tuning process in RECOMP, it is also possible to use a larger LLM that can perform summarization.
Prompt-SAW~\cite{ali2024prompt} uses a 7B Vicuna model~\cite{vicuna2023} to create a knowledge graph with the key entities and their relationships. 
Then, each entity-relation pair is encoded with an embedding model and similarity search is performed with the question embedding to determine the most relevant information to keep.
PRCA~\cite{yang2023prca} uses a small language model, such as T5-Large, to generate a smaller context from the original context. 
In order to train the small language model, PRCA uses a two-stage training process. 
In the first stage, supervised training is performed so that the small language model can learn to produce summaries well.
Then, in the second stage, proximal policy optimization is applied to train the small language to produce distilled contexts that perform well when given to the downstream LLM. 
Similarly to PCRL, the ROUGE score between the output from the policy's compressed context and the output from the original context is used to form the reward for training.

\subsubsection{Extractive Compression}
Extractive compression selects relevant documents, sentences, or phrases from the original context.
RECOMP also has an extractive compression method that is used to extract the most relevant sentences given the initial context and question.
RECOMP trains an encoder model so that useful sentences have higher inner product with the question in the embedding space. 
In their evaluation, the encoder is fine-tuned from a contriever (110M) checkpoint~\cite{izacard2021unsupervised}. 
Document rerankers perform a similar function to RECOMP's extractive compressor.
Reranker models take a question and document and output a relevance score for the document to the query.
Rerankers are typically applied in RAG pipelines after an initial retrieval step to further refine the document set. 
Prior work~\cite{nogueira2019passage} fine-tunes a BERT model~\cite{devlin2019bert} for passage rereranking.
There is also a line of work~\cite{pradeep2023rankvicuna, pradeep2023rankzephyr} that fine-tunes 7B language models to perform zero-shot listwise reranking. An illustration of extractive compression and its comparison to abstrative compression and token pruning can be found in~\autoref{fig:methods}.

\subsubsection{Query-Aware vs Query-Agnostic Compression}
Prompt compression methods may further be classified as query-aware or query-agnostic. Query-aware compression methods compress contexts differently depending on the question or task. On the other hand, query-agnostic compression methods do not rely on the question or task and thus compression may be performed offline only once. Since such methods do not have access to the downstream task, they operate by exploiting redundancy in natural language. LLMLingua-2~\cite{pan2024llmlingua2} performs query-agnostic compression by training a classifier model to identify and remove redundant tokens. Prompt-SAW also has a query-agnostic variant in which similar information elements in the constructed knowledge graph are de-duplicated. An illustration of query-aware and query-agnostic prompt compression is shown in ~\autoref{fig:query-aware-agnostic}. Furthermore,~\autoref{table:taxonomy} gives a categorization of existing methods.

\begin{table}[!t]
\caption{Existing prompt compression methods can be classified into three overarching classes: token pruning, abstractive compression, and extractive compression. Additionally, methods are distinguishable by whether or not they are query-aware.}
\vspace{-3mm}
\begin{center}
\scriptsize{
\setlength{\tabcolsep}{6pt}{
\begin{tabular}{c|c|c}
\toprule
    Class & Method & Query-Aware? \\
    \midrule
    \multirow{4}{*}{Token Pruning} & LongLLMLingua~\cite{jiang2023longllmlingua} & Yes \\
    & Attention-Based Pruning~\cite{kim2022learned} & Yes\\
    & Selective-Context~\cite{li2023compressing} & No \\
    & LLMLingua-2~\cite{pan2024llmlingua2} & No \\
    \midrule
    \multirow{2}{*}{Abstractive} & Abstractive RECOMP~\cite{xu2023recomp} & Yes \\
    & PromptSAW~\cite{ali2024prompt}  & Either \\
    \midrule
    \multirow{2}{*}{Extractive} & Extractive RECOMP~\cite{xu2023recomp} & Yes \\
    & Reranker~\cite{nogueira2019passage} & Yes \\
    \bottomrule
\end{tabular}
}
}
\end{center}
\label{table:taxonomy}
\end{table}
\section{Evaluation Methodology for Prompt Compression Methods}
\subsection{Motivation}
As shown in~\autoref{sec:related-work-prompt-compression}, there is a wide range of prompt compression techniques. 
However, there has not been a systematic study conducted on the behavior of different compression methods. 
Additionally, the evaluation schemes in existing works significantly differ.
This variation is found in benchmark selection, compression ratios, and original prompt lengths. 
LongLLMLingua primarily evaluates on prompts of size 10,000 tokens with compression ratios near $5\times$. 
On the other hand, RECOMP evaluates their extractive and abstractive compressor on much smaller prompts (500 tokens) but considers compression ratios of $20\times$. 
Prompt-SAW only evaluates their method on NaturalQuestions~\cite{kwiatkowski2019natural} and GSM8K~\cite{cobbe2021training}. 
Due to the discrepancies in evaluation methods, it is very difficult to accurately characterize the performance of prompt compression methods. 
From a practitioner's perspective, it is unclear which techniques are best applicable to their application setting. 
To resolve the lack of standardization, we perform a rigorous study of token pruning, extractive compression, and abstractive compression methods. 
There are numerous questions we aim to answer:
\begin{itemize}
    \item What are the challenges in designing effective prompt compression solutions?
    \item What are the trade-offs between different approaches to prompt compression?
    \item Are specific application settings better suited for certain methods?
\end{itemize}

\subsection{Setup}
\textbf{Models:} We use GPT-3.5-Turbo (0613 release), Mixtral 8x7B~\cite{jiang2024mixtral}, and DBRX Instruct~\cite{dbrx} as the primary LLMs.
GPT-3.5-Turbo is a proprietary model available through OpenAI, while Mixtral 8x7B and DBRX Instruct are open-source models available via Huggingface. 
All experiments are conducted with temperature zero and greedy decoding. 
Unlike Mixtral 8x7B and DBRX Instruct, GPT-3.5-Turbo is not deterministic at these settings.
Therefore, for all experiments with GPT-3.5-Turbo, we report averages over three trials. 
GPT-3.5-Turbo has a context window of 16k tokens, and both Mixtral 8x7B and DBRX Instruct have context windows of 32k tokens. 
\\ \\
\textbf{Datasets:} We conduct our evaluation using the LongBench benchmark~\cite{bai2023longbench}. 
LongBench consists of a variety of tasks that require the model to reason over large contexts of potentially tens of thousands of tokens. 
Specifically we consider three tasks that represent a wide range of popular applications: single-document question answering, multi-document question answering, and summarization. 
For each of the tasks, we consider three datasets. For single-document question answering, we use NarrativeQA~\cite{kočiský2017narrativeqa}, Qasper~\cite{dasigi2021dataset}, and MultiFieldQA-en. For multi-document question answering, we use HotpotQA~\cite{yang2018hotpotqa}, 2WikiMultihopQA~\cite{ho2020constructing}, and MuSiQue~\cite{trivedi2022musique}. 
For summarization, we use GovReport~\cite{huang2021efficient}, QMSum~\cite{zhong2021qmsum}, and MultiNews~\cite{fabbri2019multinews}. 
We use the evaluation scripts and metrics provided by LongBench.
Therefore we use F1 as the metric for question answering tasks and ROUGE~\cite{lin2004rouge} as the metric for summarization tasks.
Further descriptions of evaluated datasets, as well as their associated context lengths, may be found in~\autoref{sec:datasets}.
\\ \\
\textbf{Chunking:} In this study, chunking refers to the process of dividing the large input context into smaller, manageable segments, referred to as chunks. In our experiments, unless otherwise specified, each chunk consists of approximately 128 tokens and is carefully constructed to ensure that sentence boundaries are preserved. Chunking is crucial for methods like reranking and LongLLMLingua which operate on coarse-grained units of text by allowing each chunk to be treated as an independent document and assessed independently for its relevance to the query. The terms chunk and document are used interchangeably in our experiments.

\subsection{Evaluated Methods}
We evaluate the following prompt compression methods.

\textbf{Original:} We send the whole prompt to the LLM and truncate to the context window if necessary. 
All compression rates for other methods are reported relative to the compression rate of this method.
\\ \\
\textbf{LongLLMLingua:} We use LongLLMLingua with their suggested hyper-parameters. We vary the rate hyper-parameter to achieve different compression rates. We use a 137M GPT-2~\cite{radford2019language} as the compressor. LongLLMLingua first prunes irrelevant chunks and then performs token pruning on the kept chunks. Other hyper-parameters are set following recommended defaults, with a hyper-parameter sweep being shown in~\autoref{sec:longllmlingua-sweep}.
\\ \\
\textbf{Reranker:} We use mxbai-rerank-large-v1~\cite{mxbai} as a reranker, which is a fine-tuned DeBERTa~\cite{he2020deberta} model.
Given a question and chunk, the reranker model assigns a score from 0 to 1 denoting the relevance of the chunk to the question. The most relevant chunks are kept as context.
We vary the number of selected chunks to achieve different compression rates.
\\ \\
\textbf{Reranker + LongLLMLingua:} We replace LongLLMLingua's coarse-grained document pruning stage with a reranker model.
Then we perform token pruning with LongLLMLingua's token pruning methodology. 
We vary the rate hyper-parameter to achieve different compression rates and otherwise use the recommended hyper-parameters.
We use GPT-2 as the compressor for LongLLMLingua's token pruning method.
\\ \\
\textbf{Reranker + Token Pruning:} We implement a custom token pruning method by modifying the reranker so that it performs token-pruning while determining the relevance score for the document.
As the reranker is a DeBERTa model, we prune a fixed percentage of document tokens at each layer using attention scores. 
We prune document tokens that have the lowest attention score with respect to the query tokens. 
Our custom token pruning method compresses the initial chunk by $20\%$ by pruning $2\%$ of tokens in each of the last 10 layers.
The number of chunks selected by the reranker is varied to achieve different compression rates.
\\ \\
\textbf{Query-Agnostic Abstractive Compression:} We use Mistral 7B Instruct~\cite{jiang2023mistral} as an abstractive LLM to summarize each chunk offline.
For a user query, the reranker first selects relevant chunks and then concatenates the summaries of selected chunks to use as input for the LLM.
We ask the summarizer model to compress each chunk by $50\%$ and vary the overall compression rate by varying the number of initially selected chunks in the reranking phase. We show the summarization prompt in~\autoref{sec:abstractive-compression-prompts}.
\section{Experiments}
\label{sec:experiments}

\subsection{Main Results}
\label{sec:main-results}
\begin{figure*}[!t]
    \centering
    \includegraphics[width=\linewidth]{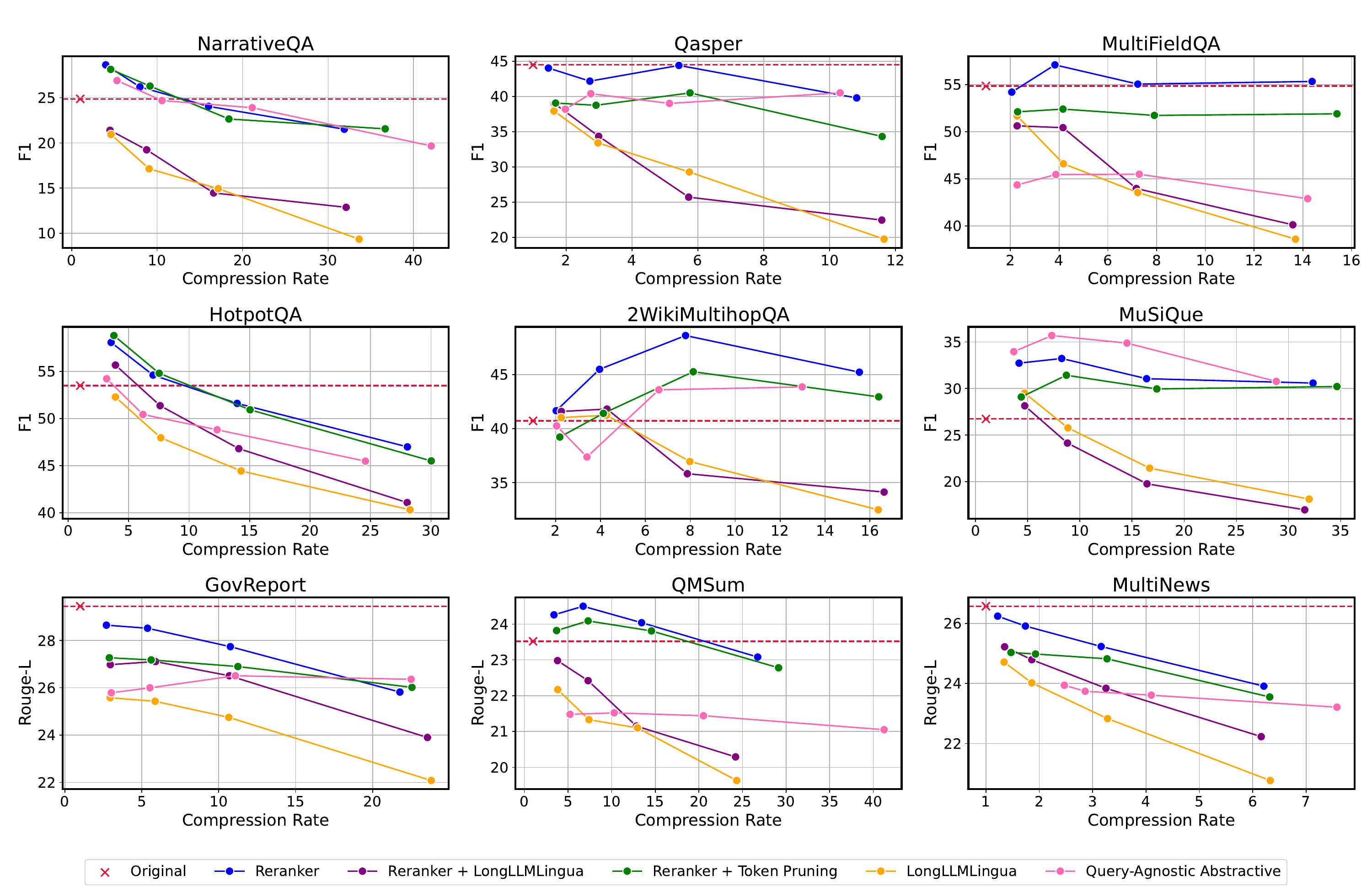}
    \caption{Results of main methods with GPT-3.5-Turbo. For each dataset, the corresponding graphs plot the accuracy metric—either F1 or Rouge-L—against the compression rate. We see similar results with Mixtral 8x7B (see~\autoref{fig:mixtral-pareto}) and DBRX Instruct (see~\autoref{fig:dbrx-pareto}).} 
    \label{fig:gpt-3.5-pareto}
\end{figure*}

The main results for GPT-3.5-Turbo are shown in~\autoref{fig:gpt-3.5-pareto}. 
We include results for Mixtral 8x7B and DBRX Instruct in~\autoref{sec:mixtral-dbrx-results} and note that it observes similar trends to GPT-3.5-Turbo.

\subsubsection{Extractive Compression}
\label{subsec:extractive_compression}
Extractive compression methods are represented via the reranker (blue). The reranker model has very strong performance across all models and datasets. 
There are many example data points where compression is performed and accuracy significantly increases. 
For example, on 2WikiMultihopQA with GPT-3.5-Turbo, the reranker is able to compress $7.75\times$ while increasing accuracy by 7.89 points. Similarly, on MuSiQue with Mixtral 8x7B, the reranker is able to compress $4.14\times$ while increasing accuracy by 7.16 points. 
On HotpotQA with Mixtral 8x7B, it is able to compress $3.55\times$ while increasing accuracy by 4.54 points. 
A significant advantage of extractive compression is that grammatical constructs are preserved, as pruning occurs at a coarse granularity.
Retrieval based methods are a widely used extractive compression methodology in which relevant chunks are retrieved via similarity search on embeddings. However, as shown in~\autoref{sec:retriever-vs-reranker} we see significant improvements in extractive compression when using a reranker model over standard retrieval. This is because reranker models use language models that take in both the query and context to assign relevance. In contrast, retrieval methods perform light-weight similarity search over embeddings.
Therefore, the precise method used for extractive compression plays a significant role.

\subsubsection{Abstractive Compression}
\label{sec:abstractive-compression-results}
Abstractive compression methods (pink) often exhibits inferior performance compared to extractive compression. 
The primary challenge with abstractive compression arises from the use of smaller, potentially weaker models, which may omit crucial information or introduce hallucinations.
This is particularly problematic in summarization tasks where the large model has to generate a summary based only on the weaker model's summaries, which can potentially discard information that the large model would have preferred to keep.
Concretely, on summarization datasets with GPT-3.5-Turbo and Mixtral 8x7B, query-agnostic abstractive compression lags behind extractive compression by 3-5 points. 
Additionally, on MultifieldQA, query-agnostic abstractive compression is typically 10-15 accuracy points below the reranker at the same compression ratio.
Therefore online query-aware abstractive compression, as shown in~\autoref{sec:query-aware-compression}, or fine-tuned summarizers may perform better than prompting out-of-the-box LMs for offline summarization.

\subsubsection{Token Pruning}
There are three token pruning methods: LongLLMLingua (orange), reranker + LongLLMLingua (purple), reranker + token pruning (green). 
We observe that LongLLMLingua and reranker + LongLLMLingua typically exhibit the worst behavior across datasets.
In~\autoref{sec:longllmlingua-sweep}, we perform a sweep over LongLLMLingua hyper-parameters but do not see any significant improvement. 
Reranker + token pruning generally trails slightly behind the plain reranker method.
We hypothesize that the lackluster performance of token pruning is due to the disruption of grammar and sentence comprehension caused by unstructured pruning.
However, we notice that reranker + token pruning outperforms the reranker model for GovReport and QMSum on Mixtral 8x7B at higher compression rates.
Similarly, on GPT-3.5-Turbo, reranker + token pruning is competitive with the plain reranker on summarization datasets at high compression rates. Nonetheless, the performance of the reranker + token pruning method trails the reranker on question-answering tasks.
In general, token pruning methods appear better suited for aggregation-style tasks that require pieces of knowledge from all segments of the initial context. 
Furthermore, rather than using out-of-the-box language models, practitioners may see better results by training language models specifically for token pruning~\cite{jung2023discrete, pan2024llmlingua2}.

\subsection{Additional Analysis}
\label{sec:additional-analysis}
 This section details our evaluations on the effects of replacing the reranker with an embedding model (\autoref{sec:retriever-vs-reranker}), performing aggressive token pruning (\autoref{sec:aggressive-pruning}), query-aware abstractive compression (\autoref{sec:query-aware-compression}), and varying chunk sizes (\autoref{sec:chunk-size}). We refer readers to the Appendix for a more comprehensive suite of additional studies on other models and datasets.
 
\subsubsection{Retriever vs Reranker}
\label{sec:retriever-vs-reranker}
\begin{figure*}[!h]
    \centering
    \includegraphics[width=\linewidth]{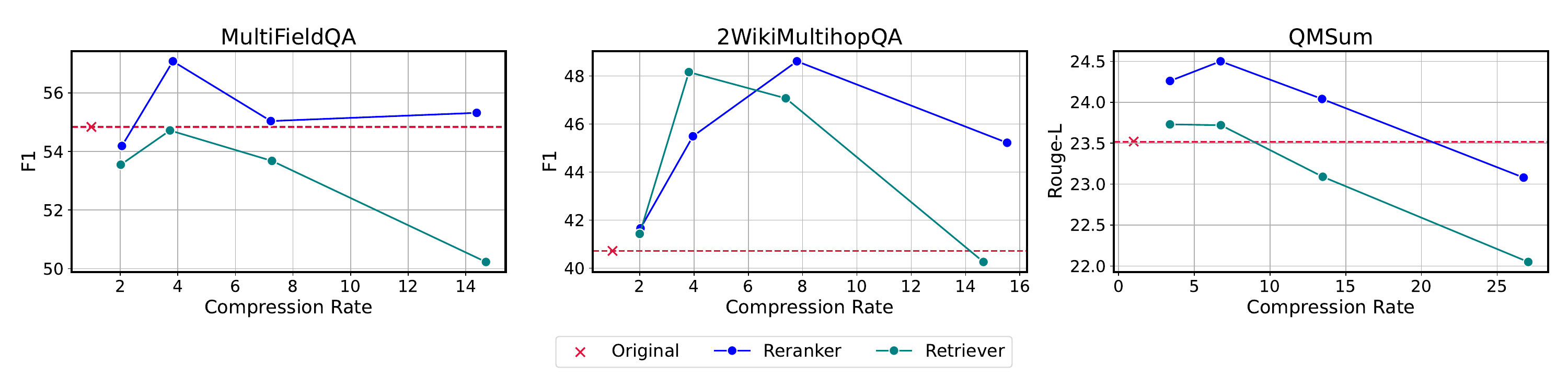}
    \caption{Analysis of performing extractive compression using standard retrieval over embedding space compared to reranking. For retrieval, embeddings are produced using text-embedding-3-small. GPT-3.5-Turbo is used as the LLM. Results on all nine datasets are shown in~\autoref{fig:retriever-gpt.3.5-full}.} 
    \label{fig:retriever-gpt.3.5}
\end{figure*}

As discussed earlier in~\autoref{subsec:extractive_compression}, instead of using a reranker for chunk-level compression, it is also possible to prune irrelevant chunks by using similarity search between the question and chunk embeddings. 
We conduct the study using OpenAI's text-embedding-3-small as the embedding model.
As shown in~\autoref{fig:retriever-gpt.3.5}, the reranker outperforms the retriever model. However, the retriever model has the advantage of requiring less resources at inference time, since document embeddings are computed offline. In many settings, reranking is applied after an initial retrieval step to reduce the number of documents that need to be reranked.

\subsubsection{Aggressive Token Pruning}
\label{sec:aggressive-pruning}
\begin{figure*}[!h]
    \centering
    \includegraphics[width=\linewidth]{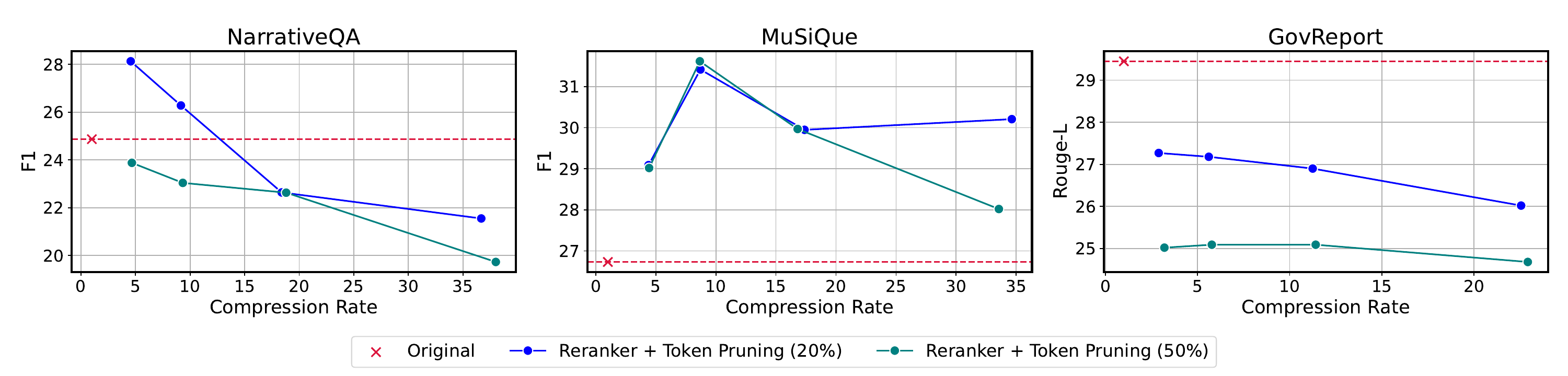}
    \caption{Performance analysis of using aggressive token pruning. We compare the original token pruning method which prunes $20\%$ of the tokens to a token pruning method that prunes $50\%$ of the tokens. When performing more aggressive token pruning, the reranker selects more chunks to achieve comparable compression ratios. GPT-3.5-Turbo is used as the LLM. Results on all nine datasets are shown in~\autoref{fig:aggresive-pruning-gpt-3.5-full}.} 
    \label{fig:aggresive-pruning-gpt-3.5}
\end{figure*}

For the token pruning methods in~\autoref{sec:main-results}, the reranker selects $25\%$ more chunks than originally and then applied a token pruning rate of $20\%$ to achieve each compression ratio. 
Here, we perform a study where the reranker selects $2\times$ more chunks and an aggressive token pruning rate of $50\%$ is applied. 
As shown in ~\autoref{fig:aggresive-pruning-gpt-3.5}, such aggressive token pruning leads to accuracy degradation. 
After observing the pruned context, we hypothesize that this is because aggressive token pruning leads to unstructured text that does not respect grammatical constructs, making it difficult for the downstream model to correctly reason over it.

\subsubsection{Query-Aware Abstractive Compression}
\label{sec:query-aware-compression}
The abstractive compression method presented in ~\autoref{sec:abstractive-compression-results} performs query-agnostic abstractive compression. 
This is largely beneficial for applications that need low-latency responses, as summaries are precomputed offline. 
However, it is also possible to perform query-aware abstractive compression, in which summaries are generated by conditioning on the question. 
Specifically, we use the reranker model to first select relevant chunks and then use a small language model to summarize the concatenation of selected chunks. We show the results in~\autoref{table:query-aware} with 16 selected chunks and include more results with 8 and 32 chunks in ~\autoref{sec:full-query-aware}. We experiment with both Mistral 7B and Llama 8B~\cite{llama3modelcard} as summarizers.
After observing difficulties in prompting such models to produce summaries of specific lengths, we used prompting methods similar to RECOMP~\cite{xu2023recomp}, which allows the Mistral model to freely choose the summarization length.
In general, our experience with abstractive compression indicates that strong prompt engineering is necessary to achieve desired performance.
The summarization prompts are shown in~\autoref{sec:abstractive-compression-prompts}. As shown in~\autoref{table:query-aware}, query-aware abstractive compression demonstrates stronger performance than query-agnostic abstractive compression. For example, on NarrativeQA, MultiFieldQA, and HotpotQA, query-aware compression performs 3-6 points better than query-agnostic. These trends persist across both Mistral 7B and Llama 8B. It is possible that more detailed summarization prompt engineering can further improve performance.
Therefore, query-aware abstractive compression may be a promising technique applications willing to handle the overhead of performing on-the-fly summarization.

\begin{table*}[!t]
\caption{Performance analysis of using query-aware abstractive compression at run time. Mistral 7B Instruct and Llama 3 8B Instruct generate summaries from chunks selected by the reranker. GPT-3.5-Turbo is used as the LLM.}
\begin{center}
\scriptsize{
\setlength{\tabcolsep}{3pt}{
\begin{tabular}{l|cc|cc|cc|cc|cc|cc|cc}
\toprule
\multirow{2}{*}{\raisebox{-0.5ex}{\textbf{Method}}} & \multicolumn{2}{c|}{\textbf{NQA}}  & \multicolumn{2}{c|}{\textbf{QAS}} & \multicolumn{2}{c|}{\textbf{MFE}}  & \multicolumn{2}{c|}{\textbf{HQA}}  & \multicolumn{2}{c|}{\textbf{WMQA}} & \multicolumn{2}{c|}{\textbf{MSQ}} & \multicolumn{2}{c}{\textbf{QMS}} \\
& Acc $\uparrow$ & CR$\uparrow$  & Acc $\uparrow$ & CR $\uparrow$ & Acc $\uparrow$ & CR $\uparrow$  & Acc $\uparrow$ & CR $\uparrow$ & Acc $\uparrow$ & CR $\uparrow$ & Acc $\uparrow$ & CR $\uparrow$  & Acc $\uparrow$ & CR $\uparrow$ \\
\midrule
\midrule

Original& 24.87 & $1.00\times$ & \textbf{44.48} & $1.00\times$ & \textbf{54.84} & $1.00\times$  & \textbf{53.50} & $1.00\times$ & 40.72 & $1.00\times$ & 26.73 & $1.00\times$  & \textbf{23.52} & $1.00\times$  \\
\midrule
\highlightrow Mistral 7B Query-Agnostic & 20.70 & $84.75\times$ & 35.63 & $20.92\times$ & 44.17 & $27.85\times$  & 48.01 & $49.56\times$ & 45.37 & $25.86\times$ & 33.27 & $57.05\times$  & 21.22 & $82.51\times$  \\
Llama 8B Query-Agnostic & 20.49 & $76.21\times$ & 33.13 & $25.11\times$ & 41.61 & $34.32\times$  & 43.51 & $60.92\times$ & 42.82 & $35.52\times$ & 28.71 & $74.18\times$  & 22.19 & $84.29\times$  \\
\midrule
\highlightrow Mistral 7B Query-Aware & \textbf{25.56} & $86.12\times$ & 36.27 & $19.96\times$ & 47.80 & $28.14\times$ & 52.23 & $44.36\times$ & \textbf{47.63} & $25.31\times$ & \textbf{33.75} & $58.28\times$ & 21.21 & $76.07\times$ \\
Llama 8B Query-Aware & 23.07 & $106.00\times$ & 38.36 & $28.71\times$ & 47.35 & $44.59\times$ & 48.81 & $91.69\times$ & 45.38 & $49.48\times$ & 28.83 & $103.24\times$ & 21.33 & $77.22\times$ \\

\bottomrule
\end{tabular}
}
}
\end{center}
\label{table:query-aware}
\end{table*}
\begin{figure*}[!t]
    \centering
    \includegraphics[width=\linewidth]{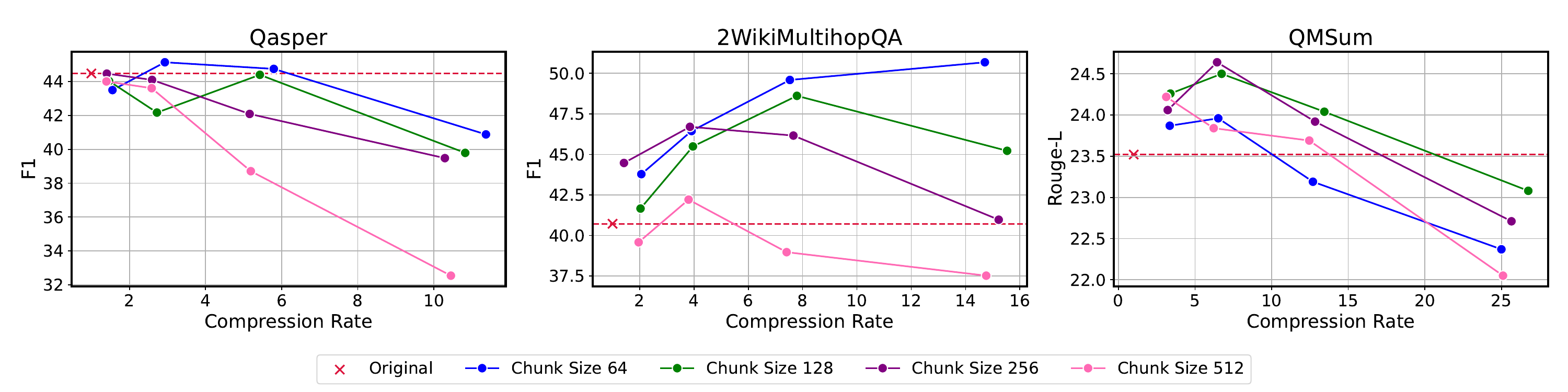}
    \caption{Impact of chunk size on the reranker with GPT-3.5-Turbo. Chunk size is varied between 64, 128, 256, and 512 tokens. Sentence boundaries are respected. Results on the token pruning reranker are shown in~\autoref{fig:chunk-size-reranker-token-pruning-gpt} and similar trends are observed.} 
    \label{fig:chunk-size-reranker-gpt}
\end{figure*}

\subsubsection{Impact of Chunk Size}
\label{sec:chunk-size}
To determine the impact of chunk size, we run a set of experiments after changing chunk size from 128 to 512 tokens.
The results are shown in~\autoref{fig:chunk-size-reranker-gpt} and~\autoref{fig:chunk-size-reranker-token-pruning-gpt}.
We notice that large chunk sizes do not perform well at large compression ratios when compared to smaller chunk sizes.
We hypothesize that this is because there are very few chunks being provided to the model when the chunk size is large. 
As a result, the model does not have the ability to see text from varying regions of the initial context.
In contrast, using smaller chunk sizes allows more chunks to be used, alleviating this issue. 
At smaller compression ratios, the chosen chunk size has lesser impact.
Ultimately, chunk size should be carefully determined after examining an application's data source as well as the desired compression ratio.
Additionally, as we demonstrate in~\autoref{sec:sql}, certain applications may require application-specific chunking techniques.

\subsection{Case Study: Text-to-SQL}
\label{sec:sql}
\begin{figure}[!t]
    \centering
    \includegraphics[width=\linewidth]{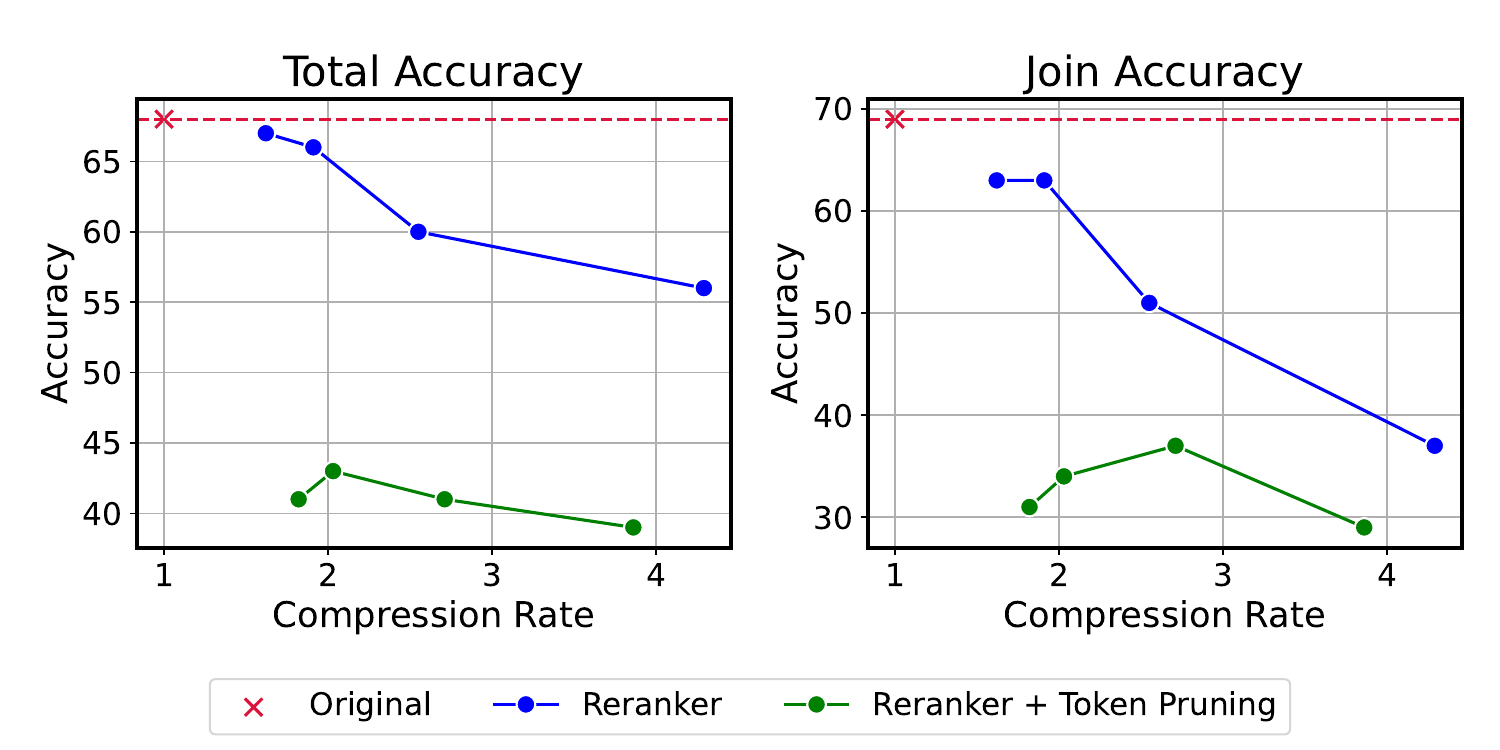}
    \caption{Results of applying the reranker and reranker + token pruning to Text-to-SQL. The total accuracy across all queries is shown in the left figure, and the accuracy across join queries is shown in the right figure. GPT-3.5-Turbo is used as the LLM.} 
    \label{fig:text-to-sql-gpt}
\end{figure}

The previous results focused on single-document, multi-document, and summarization tasks within the LongBench benchmark. 
Here, we analyze the performance of prompt compression methods when applied to Text-to-SQL.
Text-to-SQL is a popular task that requires the LLM to convert natural language into an appropriate SQL query.
We use the SQL-Eval framework~\cite{sqleval} to evaluate the impact of applying the reranker and reranker + token pruning to Text-to-SQL. In this task, CREATE TABLE statements are passed as context to the model to provide information about the different tables in the database needed to answer the question. We use the default evaluation scripts to judge accuracy, where the produced SQL query is executed within a Postgres database and compared to a ground truth. The dataset consists of 200 samples, with the context length being 1,000 tokens on average and going up to 3,000 tokens. For the reranker + token pruning method, we employ $20\%$ token pruning as in~\autoref{sec:main-results} and adjust the compression ratio by changing the number of chunks retained in the original reranking step. In order to better adhere to SQL's structure, we chunk the context so that each chunk is a single CREATE TABLE statement.
\autoref{fig:text-to-sql-gpt} shows the total accuracy across all queries, as well as the accuracy on join queries. As shown, the reranker significantly outperforms the reranker + token pruning method.
We expect that this is due to the fact that removing tokens from table definitions makes it much harder for the model to gain understanding of each table and their relation to each other. Interestingly, we notice that the join accuracy suffers significantly as the reranker's compression rate is increased. When increasing the compression rate from $1.62\times$ to $4.29\times$, the accuracy on join queries drops from 0.63 to 0.37. In contrast, the overall accuracy only drops from 0.67 to 0.56. These results are intuitive, as join queries require reasoning over multiple separate tables, which may be lost at higher compression rates.

\section{Future Directions}
There are a number of future directions to explore. As each compression method has distinct characteristics, orchestrating various methods to compress prompts is an interesting direction.
For example, LLMLingua and LongLLMLingua have a coarse-to-fine compression scheme which is a combination of extractive compression and token pruning.
Additionally, it may also be possible to develop application-specific token pruning methods that take into account the underlying nature of the context. For example, in Text-to-SQL, our general token pruning method led to significant performance loss. 
However, SQL statements have a specific grammar that may be exploited by smarter token pruning methods.
Additionally, our study focused on English datasets. It may be possible that the behavior of compression methods differs across languages which have different syntactic and semantic structures.
While our study focuses on long context inference produced by knowledge-intensive settings, it is also possible to have long prompts through many-shot prompting or verbose system prompts.
As these paradigms are different than knowledge-intensive long context inference, we leave investigation of such methods as future work.
\section{Conclusions}
This study has comprehensively characterized and evaluated a broad spectrum of existing prompt compression methods, which have become critical for long-context inference systems.
In particular, we analyze extractive compression, summarization-based abstractive compression, and token pruning methods.
Surprisingly, we find that extractive compression often outperforms all the other approaches, and enables up to $10\times$ compression with minimal accuracy degradation.
Interestingly, we also find that despite several recent claims, token pruning methods often lag behind extractive
compression. We only found marginal improvements on summarization tasks.

\section*{Acknowledgements}
We acknowledge gracious support from Furiosa and Apple team.
We also appreciate the support from Microsoft through their Accelerating Foundation Model Research, including
great support from Sean Kuno.
Furthermore, we appreciate support from
Google Cloud, the Google TRC team, and specifically Jonathan Caton, and Prof. David Patterson.
Prof. Keutzer's lab is sponsored by the Intel corporation, Intel One-API, Intel VLAB team, the Intel One-API center of
excellence, Apple, Samsung, Panasonic, as well as funding through BDD and BAIR.
Sehoon Kim  would like to acknowledge the support from the Korea Foundation for Advanced Studies (KFAS).
Amir Gholami was supported through funding from Samsung SAIT.
Our conclusions do not necessarily reflect the position or the policy of our sponsors, and no official endorsement should be~inferred.

\bibliography{ref}
\bibliographystyle{icml2024}

\renewcommand\thefigure{\thesection.\arabic{figure}} 
\setcounter{figure}{0} 

\renewcommand\thetable{\thesection.\arabic{table}} 
\setcounter{table}{0} 
\clearpage
\appendix
\clearpage
\onecolumn

\section{LongBench Dataset Details}
\label{sec:datasets}
We give a brief description of each evaluated dataset in LongBench, as well as the average token count measured by GPT-3.5-Turbo's tokenizer.
\\ \\
\textbf{NarrativeQA:} Question-answering over stories. Average tokens: 29,780.
\\ \\
\textbf{Qasper:} Question-answering over NLP papers. Average tokens: 4,923.
\\ \\
\textbf{MultiFieldQA:} Question-answering over a variety of document types such as legal documents, government reports, and academic papers. Average tokens: 6,938.
\\ \\
\textbf{HotpotQA:} 2-hop question-answering. Average tokens: 12,793.
\\ \\
\textbf{2WikiMultihopQA:} Up to 5-hop question-answering. Average tokens: 7,116.
\\ \\
\textbf{MuSiQue:} Up to 4-hop question-answering. Average tokens: 15,577.
\\ \\
\textbf{GovReport:} Summarization of detailed government reports. Average tokens: 10,242.
\\ \\
\textbf{QMSum:} Query-based summarization over meeting notes. Average tokens: 13,855.
\\ \\
\textbf{MultiNews:} Summarization of multiple news articles. Average tokens: 2,609.

\section{Additional Experimental Results}

\subsection{LongLLMLingua Hyper-Parameter Sweep}
\label{sec:longllmlingua-sweep}

\begin{figure*}[!h]
    \centering
    \includegraphics[width=\linewidth]{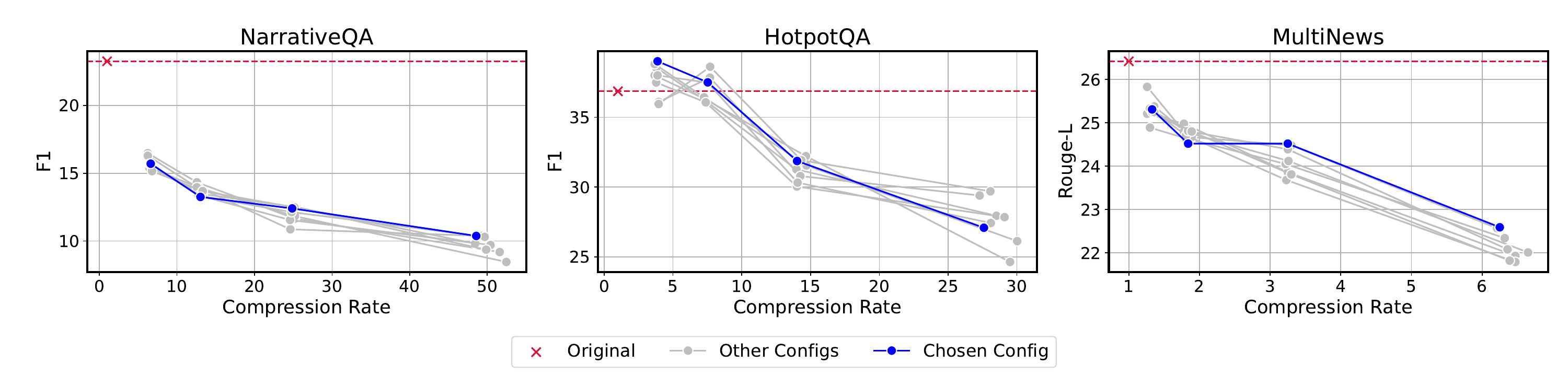}
    \caption{Analysis of performance with different LongLLMLingua hyper-parameters. The dynamic context compression ratio is varied, as well as the use of sentence-level filtering. Mixtral 8x7B is used as the LLM.} 
    \label{fig:longllmlingua-sweep-mixtral}
    \vspace{-4mm}
\end{figure*}

\begin{figure*}[!h]
    \centering
    \includegraphics[width=\linewidth]{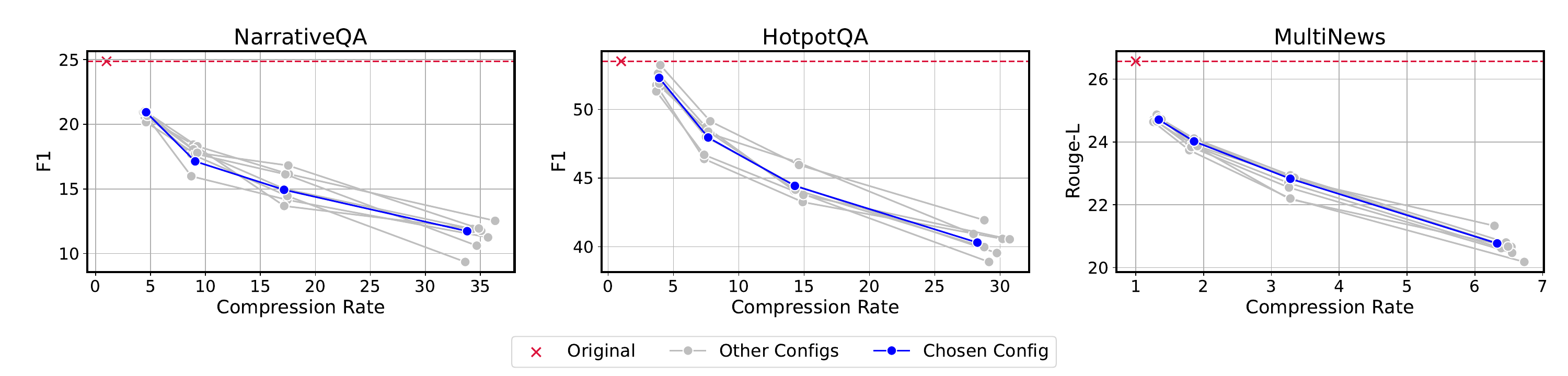}
    \caption{Analysis of performance with different LongLLMLingua hyper-parameters. The dynamic context compression ratio is varied, as well as the use of sentence-level filtering. GPT-3.5-Turbo is used as the LLM.} 
    \label{fig:longllmlingua-sweep-gpt}
    \vspace{-4mm}
\end{figure*}

In~\autoref{sec:main-results}, we used hyper-parameters for LongLLMLingua as recommended by the authors. 
Here, we perform a study where we sweep over 8 different hyper-parameter configurations for LongLLMLingua.We conduct the study on both Mixtral 8x7B and GPT-3.5-Turbo, showing the results on NarrativeQA, HotpotQA, and MultiNews. For the main results, we use the following hyper-parameters with LongLLMLingua. Sentence-level filtering turned off, dynamic context compression ratio is set to 0.3 context budget is set to $+100$, condition in question is set to ``after\_condition'',  reorder context is set to ``sort'', and condition compare is set to true. All other hyper-parameters are otherwise default. For the LongLLMLingua hyper-parameter sweep, we toggle the use of sentence-level filtering and we vary the dynamic context compression ratio between 0, 0.2, 0.3, and 0.4. As shown in~\autoref{fig:longllmlingua-sweep-mixtral} and~\autoref{fig:longllmlingua-sweep-gpt}, our chosen hyper-parameters perform well and all tested configurations exhibit similar trends.

\FloatBarrier
\subsection{Full Retriever vs Reranker Results}
\begin{figure*}[!h]
    \centering
    \includegraphics[width=\linewidth]{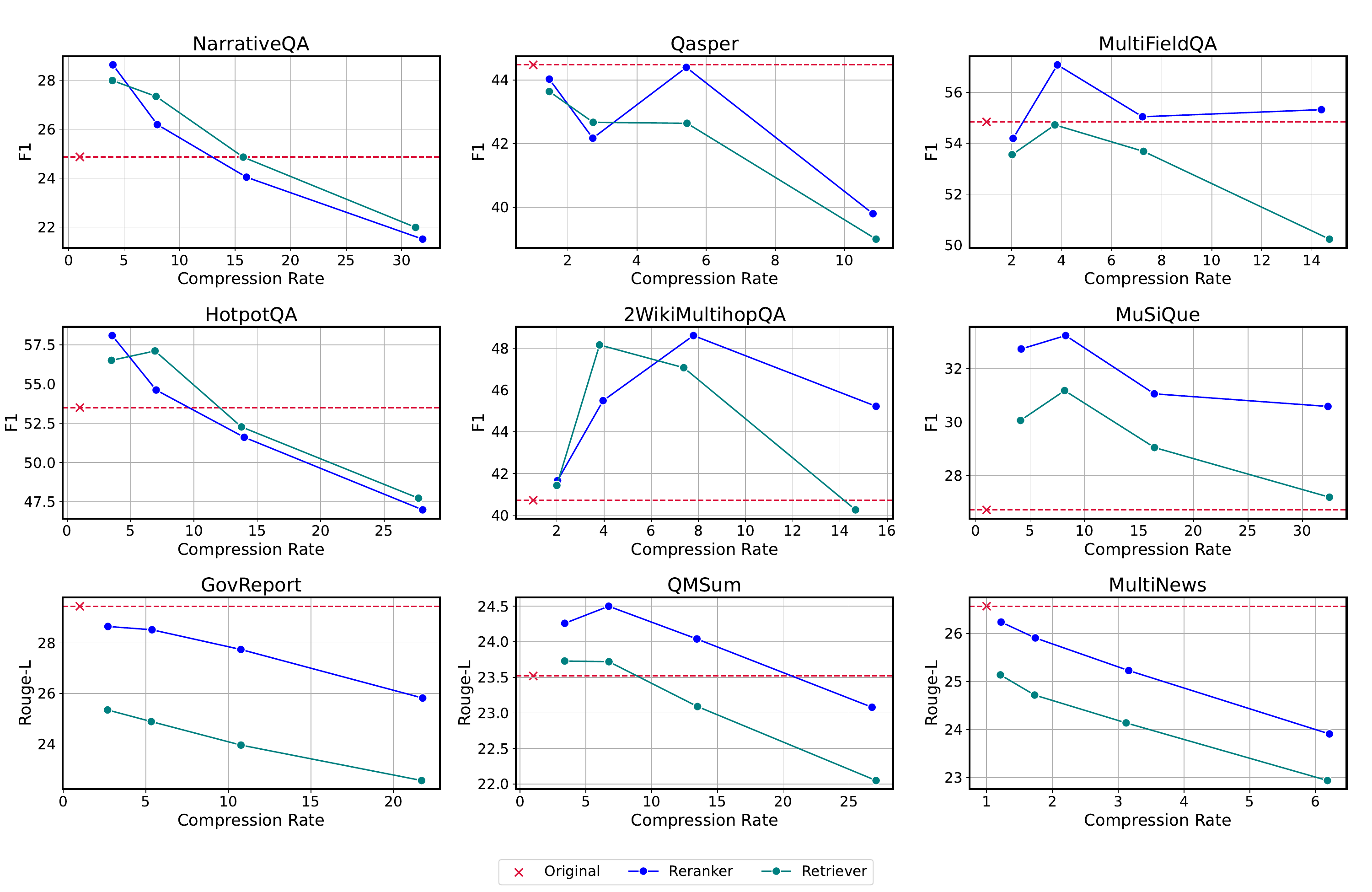}
    \caption{Analysis of performing extractive compression using standard retrieval over embedding space compared to reranking. For retrieval, embeddings are produced using text-embedding-3-small. GPT-3.5-Turbo is used as the LLM. See~\autoref{fig:retriever-gpt.3.5} for results in the main text.} 
    \label{fig:retriever-gpt.3.5-full}
\end{figure*}

In~\autoref{fig:retriever-gpt.3.5-full}, we provide the results from~\autoref{fig:retriever-gpt.3.5} on all nine datasets from LongBench.

\subsection{Full Aggressive Token Pruning Results}
\begin{figure*}[!h]
    \centering
    \includegraphics[width=\linewidth]{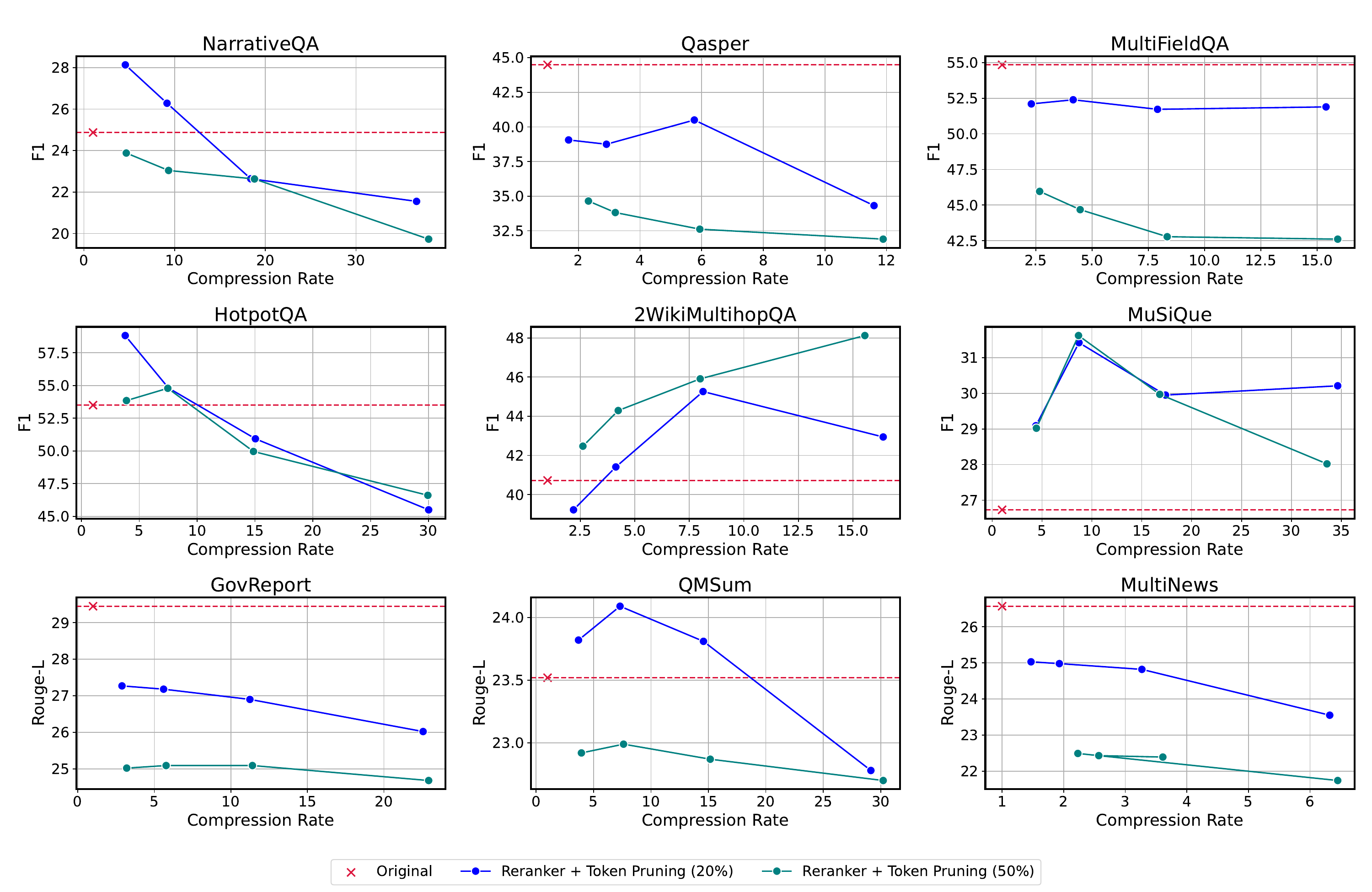}
    \caption{Performance analysis of using aggressive token pruning. We compare the original token pruning method which prunes $20\%$ of the tokens to a token pruning method that prunes $50\%$ of the tokens. GPT-3.5-Turbo is used as the LLM. See~\autoref{fig:aggresive-pruning-gpt-3.5} for results in the main text.} 
    \label{fig:aggresive-pruning-gpt-3.5-full}
\end{figure*}

In~\autoref{fig:aggresive-pruning-gpt-3.5-full}, we provide the results from~\autoref{fig:aggresive-pruning-gpt-3.5} on all nine datasets from LongBench.

\subsection{Full Query-Aware Abstractive Compression Results}
\label{sec:full-query-aware}
In~\autoref{table:full-query-aware-gpt} and~\autoref{table:full-query-aware-mixtral}, we show the results of query-aware compression on seven of the LongBench datasets, with both GPT-3.5-Turbo and Mixtral 8x7B. We also show the results with Mistral 7B and LLaMA 3 8B as summarizers. Our experiments indicate that it is difficult to control the length of summaries, making the compression rate for query-aware abstractive compression difficult to predict.
\begin{table*}[!h]
\caption{Query-aware abstractive compression results with GPT-3.5-Turbo. We use Mistral 7B Instruct and LLaMA-3 8B Instruct to generate summaries from chunks selected by the reranker. See~\autoref{table:query-aware} for results in the main text.}
\begin{center}
\scriptsize{
\setlength{\tabcolsep}{3pt}{
\begin{tabular}{l|cc|cc|cc|cc|cc|cc|cc}
\toprule
\multirow{2}{*}{\raisebox{-0.5ex}{\textbf{Method}}} & \multicolumn{2}{c|}{\textbf{NQA}}  & \multicolumn{2}{c|}{\textbf{QAS}} & \multicolumn{2}{c|}{\textbf{MFE}}  & \multicolumn{2}{c|}{\textbf{HQA}}  & \multicolumn{2}{c|}{\textbf{WMQA}} & \multicolumn{2}{c|}{\textbf{MSQ}} & \multicolumn{2}{c}{\textbf{QMS}} \\
& Acc & CR & Acc & CR & Acc & CR  & Acc & CR & Acc & CR & Acc & CR  & Acc & CR \\
\cmidrule{1-15}

\midrule
Original& 24.87 & $1.00\times$ & 44.48 & $1.00\times$ & 54.84 & $1.00\times$  & 53.5 & $1.00\times$ & 40.72 & $1.00\times$ & 26.73 & $1.00\times$  & 23.52 & $1.00\times$  \\
\midrule
\midrule
\textbf{Mistral 7B} & & & & & & \\
8 chunks & $20.48$ & $104.09\times$ & 38.36 & $21.62\times$ & 46.20 & $31.24\times$  & 49.15 & $46.19\times$ & 51.37 & $30.55\times$ & 30.71 & $65.14\times$  & 20.99 & $87.10\times$ \\
16 chunks & $25.56$ & $86.12\times$ & 36.27 & $19.96\times$ & 47.80 & $28.14\times$  & 52.23 & $44.36\times$ & 47.63 & $25.31\times$ & 33.75 & $58.28\times$  & 21.21 & $76.07\times$ \\
32 chunks & $24.12$ & $74.44\times$ & 31.70 & $19.68\times$ & 46.47 & $27.50\times$  & 50.47 & $44.47\times$ & 47.93 & $22.79\times$ & 30.49 & $52.57\times$  & 20.96 & $62.75\times$ \\
\midrule
\midrule
\textbf{LLaMA 3 8B} & & & & & & \\
8 chunks & 20.14 & $124.03\times$ & 40.86 & $35.90\times$ & 47.25 & $54.34\times$  & 48.10 & $112.60\times$ & 47.10 & $61.56\times$ & 26.56 & $124.11\times$  & 22.06 & $94.28\times$ \\
16 chunks & 23.07 & $106.00\times$ & 38.36 & $28.71\times$ & 47.35 & $44.59\times$  & 48.81 & $91.69\times$ & 45.38 & $49.48\times$ & 28.83 & $103.24\times$  & 21.33 & $77.22\times$ \\
32 chunks & 21.97 & $75.54\times$ & 33.88 & $23.89\times$ & 40.30 & $33.16\times$  & 47.18 & $64.31\times$ & 42.64 & $31.70\times$ & 30.45 & $70.56\times$  & 20.68 & $59.87\times$ \\
\bottomrule
\end{tabular}
}
}
\end{center}
\label{table:full-query-aware-gpt}
\end{table*}

\begin{table*}[!h]
\caption{Query-aware abstractive compression results with Mixtral 8x7B Instruct. We use Mistral 7B Instruct and LLaMA-3 8B Instruct to generate summaries from chunks selected by the reranker. See~\autoref{table:query-aware} for results in the main text.}
\begin{center}
\scriptsize{
\setlength{\tabcolsep}{3pt}{
\begin{tabular}{l|cc|cc|cc|cc|cc|cc|cc}
\toprule
\multirow{2}{*}{\raisebox{-0.5ex}{\textbf{Method}}} & \multicolumn{2}{c|}{\textbf{NQA}}  & \multicolumn{2}{c|}{\textbf{QAS}} & \multicolumn{2}{c|}{\textbf{MFE}}  & \multicolumn{2}{c|}{\textbf{HQA}}  & \multicolumn{2}{c|}{\textbf{WMQA}} & \multicolumn{2}{c|}{\textbf{MSQ}} & \multicolumn{2}{c}{\textbf{QMS}} \\
& Acc & CR & Acc & CR & Acc & CR  & Acc & CR & Acc & CR & Acc & CR  & Acc & CR \\
\cmidrule{1-15}

\midrule
Original& 23.26 & $1.00\times$ & 31.66 & $1.00\times$ & 47.36 & $1.00\times$  & 36.86 & $1.00\times$ & 26.51 & $1.00\times$ & 18.11 & $1.00\times$  & 24.92 & $1.00\times$  \\
\midrule
\midrule
\textbf{Mistral 7B} & & & & & & \\
8 chunks & 15.65 & $165.39\times$ & 25.65 & $21.25\times$ & 42.29 & $32.26\times$ & 38.01 & $53.95\times$ & 34.75 & $31.32\times$ & 19.81 & $66.88\times$ & 21.88 & $103.89\times$ \\
16 chunks & 15.34 & $135.35\times$ & 23.62 & $19.68\times$ & 44.82 & $28.25\times$ & 38.11 & $47.34\times$ & 29.79 & $27.27\times$ & 21.53 & $59.21\times$ & 21.16 & $90.03\times$ \\
32 chunks & 18.17 & $118.15\times$ & 19.86 & $20.21\times$ & 40.86 & $28.00\times$ & 39.84 & $44.85\times$ & 30.82 & $26.75\times$ & 19.68 & $55.93\times$ & 21.08 & $78.37\times$ \\
\midrule
\midrule
\textbf{LLaMA 3 8B} & & & & & & \\
8 chunks & 14.13 & $197.32\times$ & 25.82 & $35.58\times$ & 41.72 & $53.79\times$ & 34.83 & $116.14\times$ & 28.61 & $63.79\times$ & 16.17 & $135.038\times$ & 22.06 & $112.08\times$ \\
16 chunks & 6.21 & $167.87\times$ & 24.53 & $28.64\times$ & 42.90 & $45.19\times$ & 39.20 & $96.06\times$ & 27.93 & $50.03\times$ & 20.99 & $106.54\times$ & 21.14 & $91.13\times$ \\
32 chunks & 17.66 & $118.51\times$ & 21.87 & $24.10\times$ & 37.80 & $33.52\times$ & 33.35 & $67.31\times$ & 25.05 & $33.62\times$ & 15.84 & $75.27\times$ & 21.40 & $70.54\times$ \\
\bottomrule
\end{tabular}
}
}
\end{center}
\label{table:full-query-aware-mixtral}
\end{table*}

\subsection{Abstractive Compression Prompts}
In~\autoref{table:summarization-prompts}, we show the prompts used to perform query-aware and query-agnostic abstractive compression.
\label{sec:abstractive-compression-prompts}
\begin{table*}[!h]
\vspace{-3mm}
\caption{Prompts used for query-aware and query-agnostic abstractive compression.}
\vspace{-3mm}
\begin{center}
\scriptsize{
\setlength{\tabcolsep}{6pt}{
\begin{tabular}{l|l}
\toprule
    Method & Prompt \\
    \midrule
    \multirow{2}{*}{Query-Agnostic} & Could you please rephrase the paragraph to make it short, and keep $50\%$ tokens. \\
    & Respond with ONLY the compressed paragraph and nothing else. Paragraph: \texttt{paragraph} \\
    \midrule
    \multirow{2}{*}{Query-Aware (Mistral 7B Instruct)} & Compress the information in the retrieved documents into a summary that could be \\
    & used to answer the question: Question: \texttt{query} Retrieved documents: \texttt{docs} \\
    \midrule
    \multirow{2}{*}{Query-Aware (LLaMA 3 8B Instruct)} & Compress the information in the retrieved documents into a summary that could be used to answer the question.\\
    & Do NOT try to directly answer the question. Question: \texttt{query} Retrieved documents: \texttt{docs} \\
    \bottomrule
\end{tabular}
}
}
\end{center}
\label{table:summarization-prompts}
\end{table*}
\subsection{Impact of Weaker Reranker Model}
\label{sec:weak-reranker-model}
\begin{figure*}[!h]
    \centering
    \includegraphics[width=\linewidth]{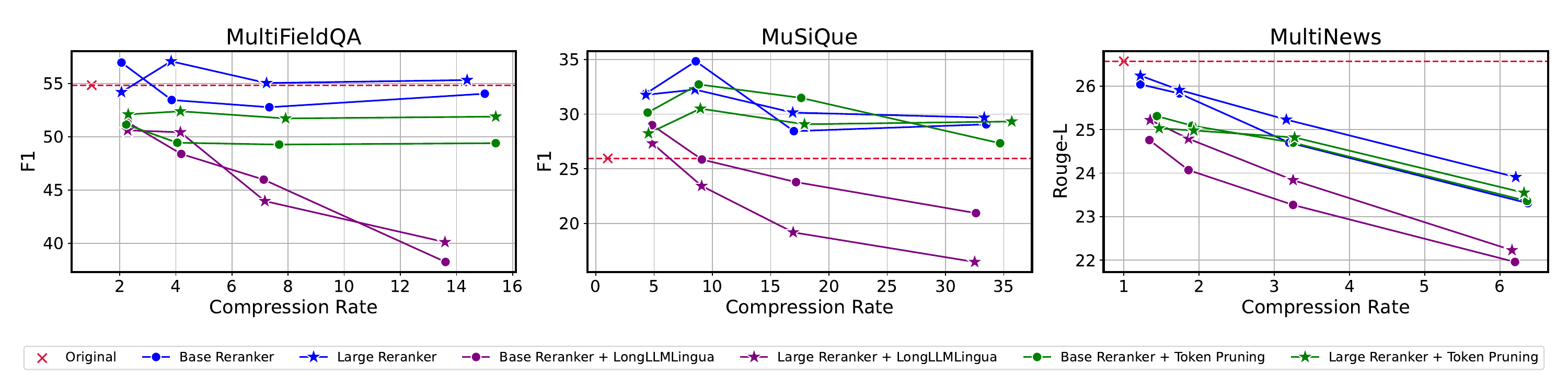}
    \caption{Performance comparison between using mxbai-rerank-large-v1 (435M) versus mxbai-rerank-base-v1 (184M) with GPT-3.5-Turbo as the LLM.} 
    \label{fig:weak-reranker-gpt-3.5}
\end{figure*}

\begin{figure*}[!h]
    \centering
    \includegraphics[width=\linewidth]{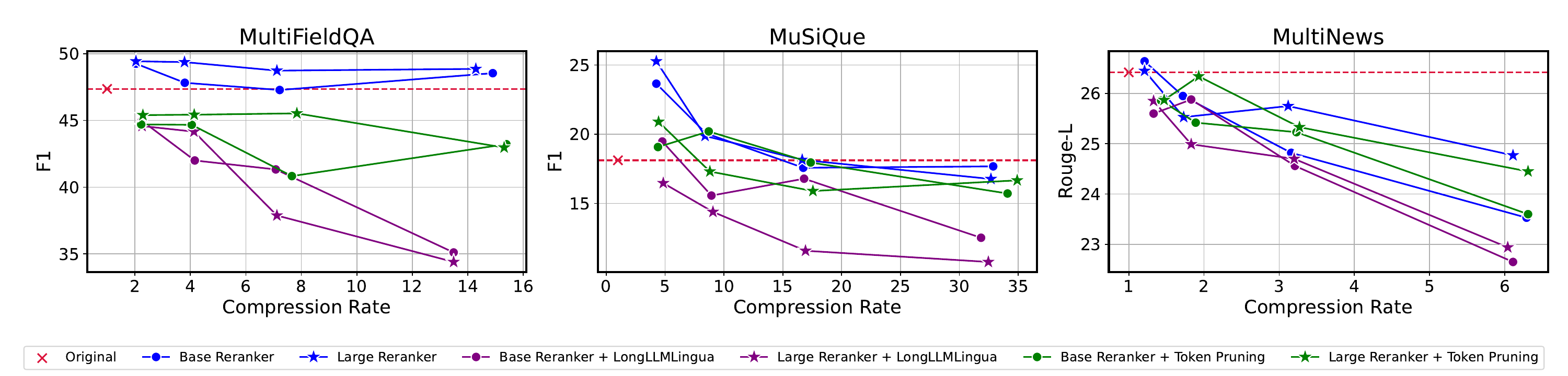}
    \caption{Performance comparison between using mxbai-rerank-large-v1 (435M) versus mxbai-rerank-base-v1 (184M) with Mixtral 8x7B as the LLM.} 
    \label{fig:weak-reranker-mixtral}
\end{figure*}

In~\autoref{sec:main-results}, we used mxbai-rerank-large-v1 (435M) as the reranker. Here, we perform a study when using a weaker reranker model, namely mxbai-rerank-base-v1 (184M), as certain applications may have access to lesser computing resources or have stronger latency requirements.
Since mxbai-rerank-base-v1 only has 12 layers, we modify our custom token pruning scheme to prune by $4\%$ starting from layer 8. 
As shown in~\autoref{fig:weak-reranker-gpt-3.5} and~\autoref{fig:weak-reranker-mixtral}, the large reranker generally outperforms the base reranker across all three datasets. 
However, there are certain points at which the base reranker outperforms the large reranker. 
Thus the base reranker can be a suitable alternative in resource constrained settings.
\FloatBarrier
\subsection{Impact of Chunk Size on Token Pruning Reranker}
\begin{figure*}[!h]
    \centering
    \includegraphics[width=\linewidth]{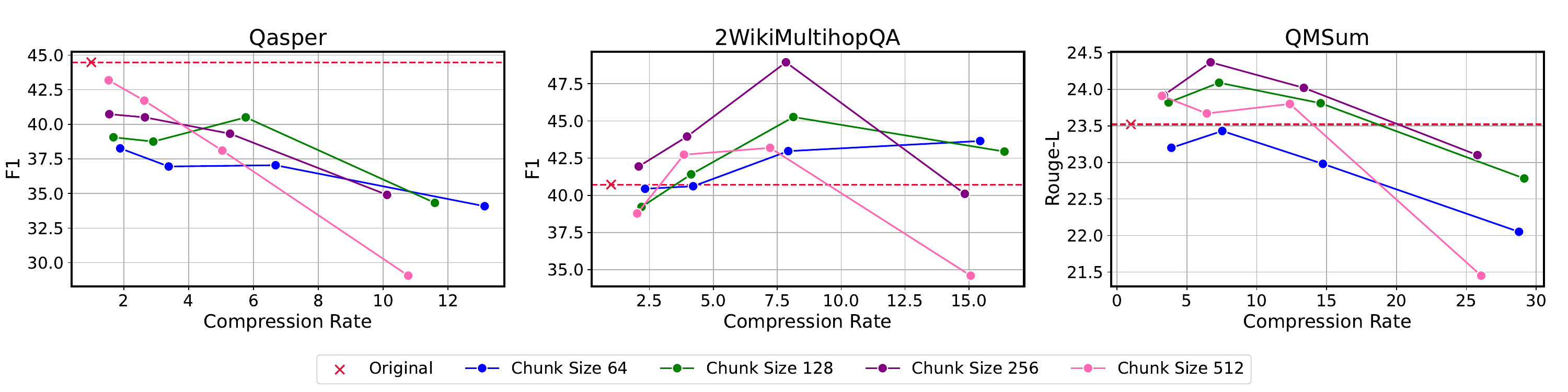}
    \caption{Impact of chunk size on the token pruning reranker with GPT-3.5-Turbo. Chunk size is varied between 64, 128, 256, and 512 tokens. Sentence boundaries are respected. See~\autoref{fig:chunk-size-reranker-gpt} for impact of chunk size on the reranker.} 
    \label{fig:chunk-size-reranker-token-pruning-gpt}
\end{figure*}

In~\autoref{fig:chunk-size-reranker-token-pruning-gpt}, we show the impact of chunk size on the token pruning reranker.
\newpage
\subsection{Mixtral 8x7B and DBRX Instruct Results}
\label{sec:mixtral-dbrx-results}
\begin{figure*}[!h]
    \centering
    \includegraphics[width=0.8\linewidth]{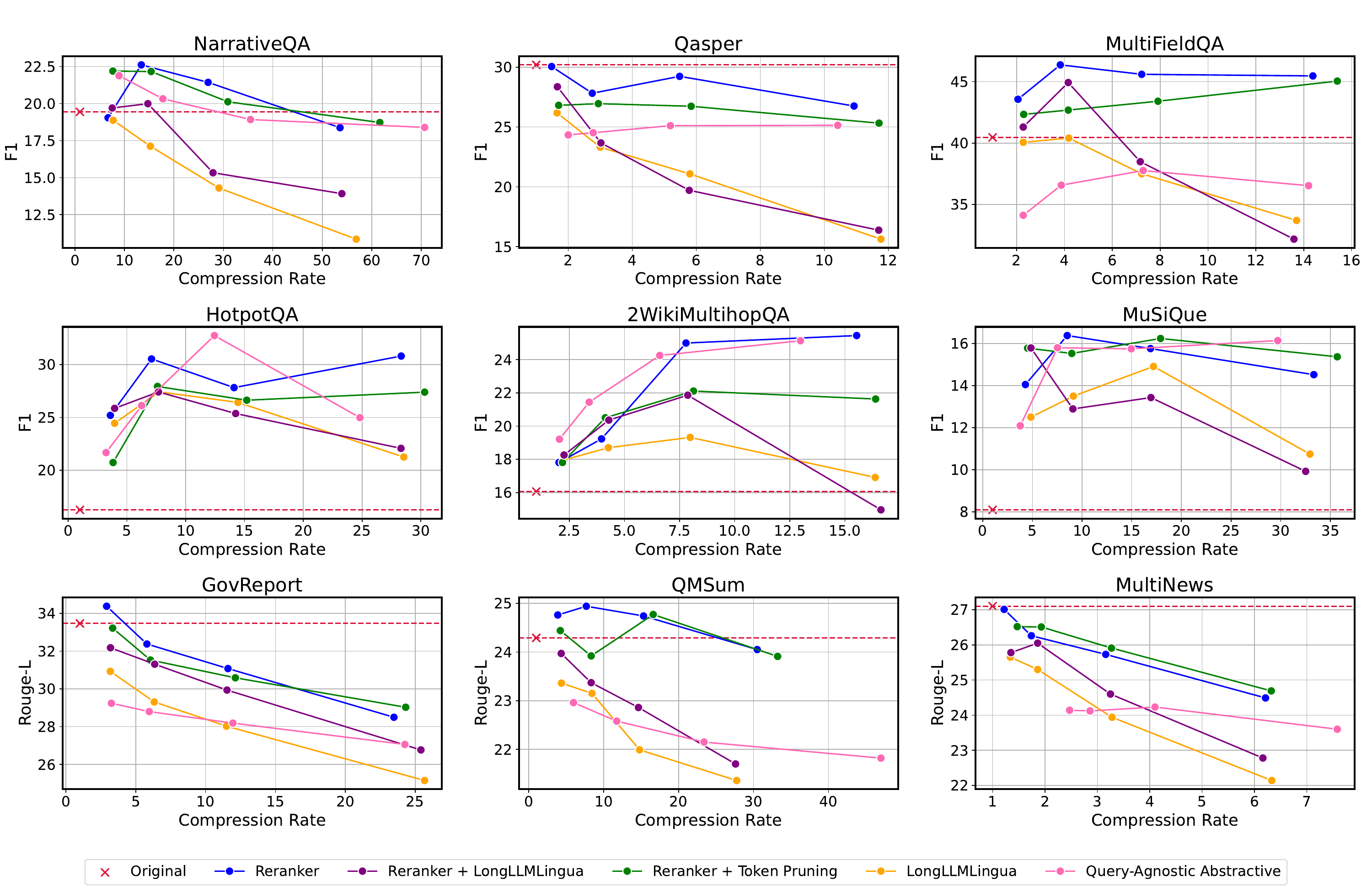}
    \caption{Results of main methods with DBRX Instruct. For each dataset, the corresponding graphs plot the accuracy metric—either F1 or Rouge-L—against the compression rate. See~\autoref{fig:gpt-3.5-pareto} for results on GPT-3.5-Turbo.} 
    \label{fig:dbrx-pareto}
\end{figure*}

\begin{figure*}[!h]
    \centering
    \includegraphics[width=0.8\linewidth]{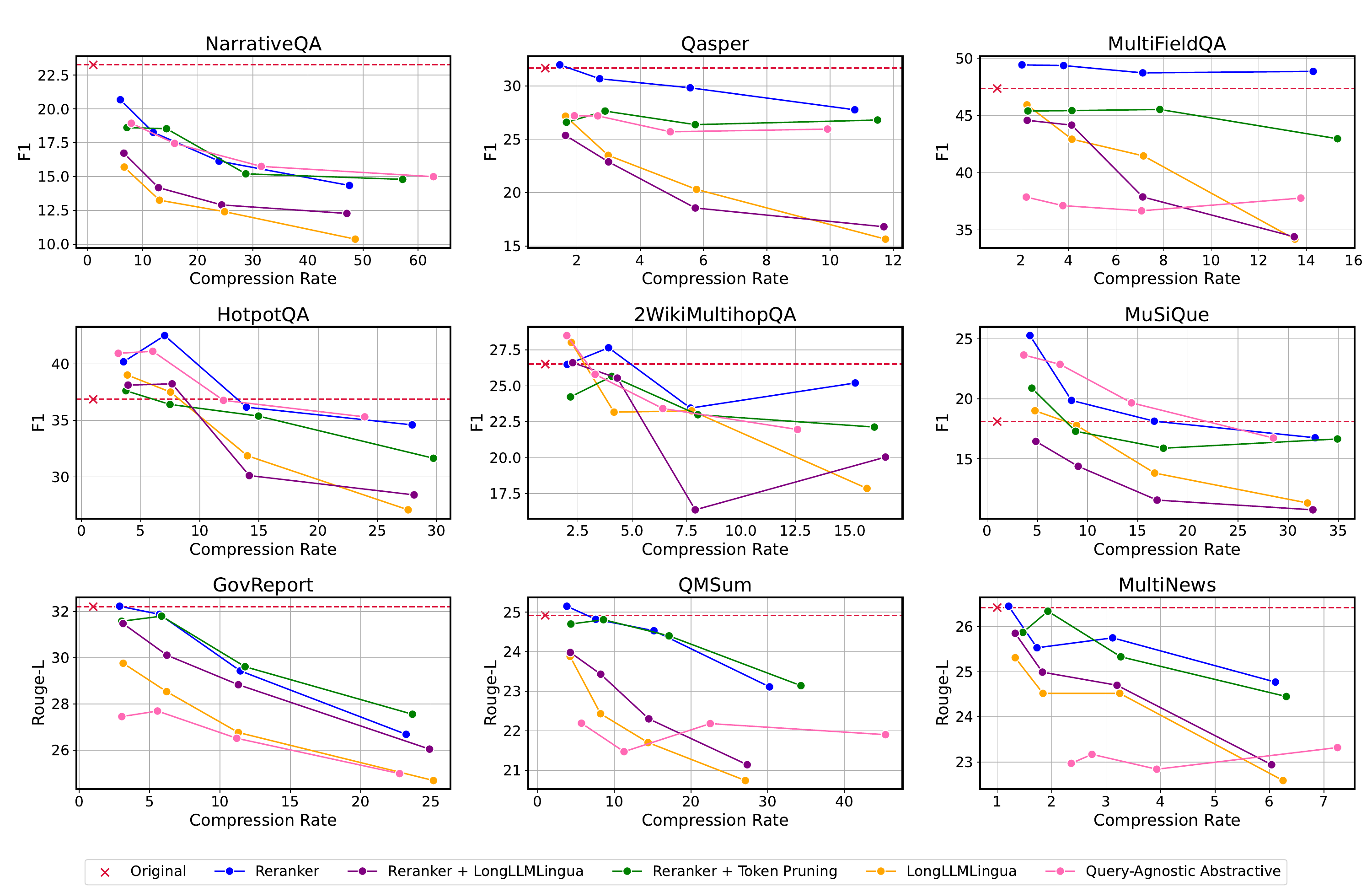}
    \caption{Results of main methods with Mixtral 8x7B.  For each dataset, the corresponding graphs plot the accuracy metric—either F1 or Rouge-L—against the compression rate. See~\autoref{fig:gpt-3.5-pareto} for results on GPT-3.5-Turbo.} 
    \label{fig:mixtral-pareto}
\end{figure*}

In~\autoref{fig:mixtral-pareto} and~\autoref{fig:dbrx-pareto}, we show the results of various compression methods on Mixtral 8x7B and DBRX Instruct.

\end{document}